\documentclass[10pt,twocolumn,letterpaper]{article}

\usepackage{wacv}
\usepackage{times}
\usepackage{epsfig}
\usepackage{graphicx}
\usepackage{amsmath}
\usepackage{amssymb}
\usepackage{times}
\usepackage{amsmath}
\usepackage{amssymb}
\usepackage{array}
\usepackage{url}
\usepackage{comment}
\usepackage{rotating}
\usepackage{multirow}
\usepackage{subfigure}
\usepackage{color}
\usepackage{acronym}
\usepackage{algorithmic}
\usepackage[normalem]{ulem}
\usepackage{refcount}
\usepackage[pagebackref=true,breaklinks=true,letterpaper=true,colorlinks,bookmarks=false]{hyperref}
\usepackage{enumitem}
\usepackage[toc,page]{appendix}

\newcommand{\mI}{{\boldsymbol{I}}}
\newcommand{\mIh}{{\widehat{\boldsymbol{I}}}}
\newcommand{\mTheta}{{\boldsymbol{\Theta}}}
\newcommand{\mPhi}{{\boldsymbol{\Phi}}}
\newcommand{\vpsi}{{\boldsymbol{\psi}}}
\newcommand{\mG}{{\boldsymbol{G}}}
\newcommand{\mD}{{\boldsymbol{D}}}

\wacvfinalcopy 

\def\wacvPaperID{342} 

\begin{document}


\title{Identity-preserving Face Recovery from Portraits}


\author{Fatemeh~Shiri\textsuperscript{1}, Xin~Yu\textsuperscript{1}, Fatih~Porikli\textsuperscript{1}, Richard~Hartley\textsuperscript{1,2}, Piotr~Koniusz\textsuperscript{2,1}\\
$^1\!$Australian National University, $^2$Data61/CSIRO\\
{\tt\small firstname.lastname@\{anu.edu.au\textsuperscript{1}, data61.csiro.au\textsuperscript{2}\}}
}

\maketitle
\ifwacvfinal\thispagestyle{empty}\fi

\begin{abstract}

Recovering the latent photorealistic faces from their artistic portraits aids human perception and facial analysis. 
However, 
a recovery process
that can preserve identity is challenging because the fine details of real faces can be distorted or lost in stylized images.
In this paper, we present a new Identity-preserving Face Recovery from Portraits (IFRP) to recover latent photorealistic faces from unaligned stylized portraits.
Our IFRP method consists of two components: Style Removal Network (SRN) and Discriminative Network (DN). 
The SRN is designed to transfer feature maps of stylized images to the feature maps of the corresponding photorealistic faces.
By embedding spatial transformer networks into the SRN, our method can compensate for misalignments of stylized faces automatically and output aligned realistic face images. The role of the DN is to enforce recovered faces to be similar to authentic faces.
To ensure the identity preservation, we promote the recovered and ground-truth faces to share similar visual features via a distance measure which compares features of recovered and ground-truth faces extracted from a pre-trained VGG network. 
We evaluate our method on a large-scale synthesized dataset of real and stylized face pairs and attain state of the art results. 
In addition, our method can 
recover photorealistic faces from previously unseen stylized portraits, original paintings and human-drawn sketches.
\end{abstract}

\renewcommand{\thefootnote}{\fnsymbol{footnote}}
\footnotetext[1]{\label{foot:maps}This work has been published in WACV'18.\vspace{-0.6cm}}

\vspace{-0.0cm}
\section{Introduction}
\label{sec:introduction}
\begin{figure}[t]
\vspace{-0.3cm}
\begin{minipage}{0.18\linewidth}
\centering
\subfigure[Original]{\label{fig:opena}\scalebox{1}[1]{\includegraphics[width=1\linewidth]{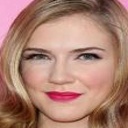}}}
\end{minipage}
\hspace{-0.5em}
\begin{minipage}{0.82\linewidth}
\centering
\subfigure[Seen]{\label{fig:openb}\scalebox{1}[1]{\includegraphics[width=0.234\linewidth]{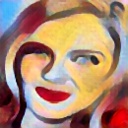}}}
\subfigure[\scriptsize{Landmarks}]{\label{fig:openc}\scalebox{1}[1]{\includegraphics[width=0.234\linewidth]{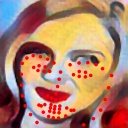}}}
\subfigure[\cite{johnson2016perceptual}]{\label{fig:opend}\scalebox{1}[1]{\includegraphics[width=0.234\linewidth]{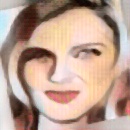}}}
\subfigure[Ours]{\label{fig:opene}\scalebox{1}[1]{\includegraphics[width=0.234\linewidth]{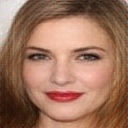}}}\\
\vspace{-0.8em}
\hspace{0.01em}
\subfigure[Unseen]{\label{fig:openf}\scalebox{1}[1]{\includegraphics[width=0.234\linewidth]{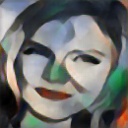}}}
\subfigure[\scriptsize{Landmarks}]{\label{fig:openg}\scalebox{1}[1]{\includegraphics[width=0.234\linewidth]{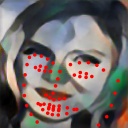}}}
\subfigure[\cite{johnson2016perceptual}]{\label{fig:openh}\scalebox{1}[1]{\includegraphics[width=0.234\linewidth]{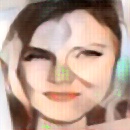}}}
\subfigure[Ours]{\label{fig:openi}\scalebox{1}[1]{\includegraphics[width=0.234\linewidth]{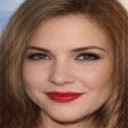}}}
\vspace{-0.4em}
\end{minipage}
\vspace{0.05cm}
\caption{Comparisons to the state-of-art method. (a) Ground-truth face image (from test dataset; not available in the training dataset). (b, f) Unaligned stylized portraits of (a) from \emph{Candy} style (seen/used style in training) and \emph{Udnie} style (unseen style in training), respectively. (c, g) Detected landmarks by~\cite{zhang2014facial}. (d, h) Results obtained by \cite{johnson2016perceptual}. (e, i) Our results.} 
\label{fig:open}
\vspace{-0.4cm}
\end{figure}

A variety of style transfer methods have been proposed to generate portraits in different artistic styles from photorealistic images. However, the recovery of photorealistic faces from artistic portraits has not been fully investigated yet. 
In general, stylized face images contain various facial expressions, facial component distortions and misalignments. Therefore, landmark detectors often fail to localize facial landmarks accurately as shown in Figures \ref{fig:openc} and \ref{fig:openg}. Thus, restoring identity-consistent photorealistic face images from unaligned stylized ones is challenging.   

While recovering photorealistic images from portraits is still uncommon in the literature, image stylization methods have been widely studied. Recently, 
Gatys~\emph{et al.}~\cite{gatys2016controlling} achieve promising results by transferring different styles of artworks to images via the semantic contents space. Since this method generates the stylized images by iteratively updating the feature maps of CNNs, it requires costly computations. In order to reduce the computational complexity, several feed-forward CNN based methods have been proposed \cite{ulyanov2016texture,ulyanov2016instance,johnson2016perceptual,dumoulin2016,li2017diversified,chen2016fast,zhang2017multi,huang2017arbitrary}. 
However, these methods can use only a single style fixed during the training phase. Such methods are insufficient for generating photorealistic face images, as shown in Figures~\ref{fig:opend} and \ref{fig:openh}, because they only capture the correlations of feature maps by the use of Gram matrices and discard spatial relations \cite{pk_tensor,pk_da,sparse_tensor_cvpr}.

In order to capture spatially localized statistics of a style image, several patch-based methods~\cite{li2016precomputed,isola2016image} have been developed. However, such methods cannot capture the global structure of faces either, thus failing to generate authentic face images. For instance, patch-based methods \cite{li2016precomputed,isola2016image} fail to attain consistency of face colors, as shown in Figure~\ref{fig:cmp2e}. Furthermore, the state-of-the-art style transfer methods~\cite{gatys2016controlling,li2016precomputed,ulyanov2016texture,johnson2016perceptual} transfer the desired styles to the given images without considering the task of identity preservation. Hence, previous methods cannot generate real faces while preserving identity.

In this paper, we develop a novel end-to-end trainable identity-preserving approach to face recovery that automatically maps the unaligned stylized portraits to aligned photorealistic face images. 
Our network employs two subnetworks: a generative subnetwork, dubbed Style Removal Network (SRN), and a Discriminative Network (DN). 
The SRN consists of an autoencoder (a downsampling encoder and an upsampling decoder) and Spatial Transfer Networks (STN)~\cite{jaderberg2015spatial}.
Th encoder extracts facial components from unaligned stylized face images and transfer the extracted feature maps to the domain of photorealistic images. Subsequently, 
our decoder forms face images. STN layers are used by the encoder and decoder to align stylized faces. 
The discriminative network, inspired by~\cite{Goodfellow2014,denton2015deep,yu2016ultra,yu2017face}, forces SRN to generate destylized faces to be similar to authentic ground-truth faces.

Moreover, as we aim to preserve the facial identity information, we constrain the recovered faces to have the same CNN feature representations as the ground-truth real faces. For this purpose, we employ pixel-level Euclidean and identity-preserving loss functions to guarantee the appearance- and identity-wise similarity to the ground-truth data. We also use an adversarial loss to achieve  high-quality visual results.   

To train our network, we require pairs of Stylized Face (SF) and ground-truth Real Face (RF) images. Therefore, we synthesize a large-scale dataset of SF/RF pairs. We observe that our CNN filters learned on images of seen styles (used for training) can extract meaningful features from images in unseen styles. Thus, the facial information of unseen stylized portraits can be extracted and used to generate photorealistic faces, as shown in the experimental section. 


The main contributions of our work are fourfold:
\setlist[enumerate,1]{label={(\roman*)}}
\vspace{0.2cm}
\begin{enumerate}[topsep=0pt,itemsep=0.1pt,leftmargin=16pt] 
\item We propose an IFRP approach that can recover photorealistic faces from unaligned stylized portraits. Our method generates facial identities and expressions that match the ground-truth face images well. 

\item We use STNs as intermediate layers to compensate for misalignments of input portraits. Thus, our method does not require the use of facial landmarks or 3D face models (typically used for face alignment).

\item We fuse an identity-preserving loss, a pixel-wise similarity loss and an adversarial loss to remove seen/unseen styles from portraits and recover the underlying identity. 

\item As large-scale datasets of stylized and photorealistic face pairs are not available, we synthesize a large dataset of pairs of stylized and photorealistic faces, which will be available on-line.

\end{enumerate}\vspace{0.2cm}

To the best of our knowledge, our method is the first attempt to provide a unified approach to the automated style removal of unaligned stylized portraits.

\section{Related Work}
\label{sec:Related Work}
In this section, we briefly review neural generative models and deep style transfer methods for image generation.

\subsection{Neural Generative Models}
There exist many generative models for the problem of image generation~\cite{oord2016pixel,kingma2013auto,oord2016pixel,Goodfellow2014,denton2015deep,zhang2017image,Shiri2017FaceD}. Among them, GANs are conceptually closely related to our problem as they employ an adversarial loss that forces the generated images to be as photorealistic as the ground-truth images. 

Several methods adopt an adversarial training to learn a parametric translating function from a large-scale dataset of input-output pairs, such as super-resolution~\cite{ledig2016photo,yu2017face,huang2017beyond,yu2017hallucinating,yu2016ultra} and inpainting~\cite{pathak2016context}. These approaches often use the $\ell_2$ or $\ell_1$ norm and adversarial losses to compare the generated image to the corresponding ground truth image. Although these methods produce impressive photorealistic images, they fail to preserve identities of subjects.


Conditional GANs have been used for the task of generating photographs from sketches \cite{sangkloy2016scribbler}, and from semantic layout and scene attributes \cite{karacan2016learning}. Li and Wand \cite{li2016precomputed} train a Markovian GAN for the style transfer -- a discriminative training is applied on Markovian neural patches to capture local style statistics. 
Isola \emph{et al.} \cite{isola2016image} develop ``pix2pix'' framework which uses so-called ``Unet'' architecture and the patch-GAN to transfer low-level features from the input to the output domain. For faces, this approach produces visual artefacts and fails to capture the global structure of faces.


Patch-based methods fail to capture the global structure of faces and, as a result, they generate poor destylization results. 
In contrast, we propose an identity-preserving loss to faithfully recover the most prominent details of faces.


Moreover, there exist several methods to synthesize sketches from photographs (and vice versa) \cite{nejati2011study,yuen2007human,tang2003face,sharma2011bypassing}. While sketch-to-face synthesis is a related problem, our unified framework can work with various more complex styles.

\subsection{Deep Style Transfer}
Style transfer is a technique which can render a given content image (input) by incorporating a specific painting style while preserving the contents of input. 
We distinguish \emph{image optimization-based} and \emph{feed-forward} style transfer methods. The seminal optimization-based work~\cite{gatys2016image} transfers the style of an artistic image to a given photograph. It uses an iterative optimization to generate a target image which is randomly initialized (Gaussian distribution). During the optimization step, the statistics of the neural activations of the target, the content and style images are matched.
%

The idea~\cite{gatys2016image} inspired many follow-up studies. 
Yin \cite{yin2016content} presents a content-aware style transfer method which initializes the optimization algorithm with a content image instead of a random noise. Li and Wand \cite{li2016combining} propose a patch-based style transfer method by combining Markov Random Field (MRF) and CNN techniques. The work \cite{gatys2016preserving} proposes to transfer the style by using linear models. It preserves colors of content images by matching color histograms. 

Gatys ~\emph{et al.}~\cite{gatys2016controlling} decompose styles into perceptual factors and then manipulate them for the style transfer. Selim~\emph{et al.} \cite{selim2016painting} modify the content loss through a gain map for the head portrait painting transfer. 
Wilmot ~\emph{et al.}~\cite{wilmot2017stable} use histogram-based losses in their objective and build on the Gatys~\emph{et al.}'s algorithm~\cite{gatys2016image}. Although the above optimization-based methods further improve the quality of style transfer, they are computationally expensive due to the iterative optimization procedure, thus limiting their practical use.

To address the poor computational speed, feed-forward methods replace the original on-line iterative optimization step with training a feed-forward neural network off-line and generating stylized images on-line~\cite{ulyanov2016texture,johnson2016perceptual,li2016precomputed}.

Johnson \emph{et al.} \cite{johnson2016perceptual} train a generative network for a fast style transfer using perceptual loss functions. 
The architecture of their generator network follows the work \cite{radford2015unsupervised} and also uses residual blocks. Another concurrent work~\cite{ulyanov2016texture}, named Texture Network, employs a multi-resolution architecture in the generator network. 
Ulyanov \emph{et al.} \cite{ulyanov2016instance,ulyanov2017improved} replace the spatial batch normalization with the instance normalization to achieve a faster convergence. Wang \emph{et al.}~\cite{wang2016multimodal} enhance the granularity of the feed-forward style transfer with multimodal CNN which performs stylization hierarchically via multiple losses deployed across multiple scales. 


These feed-forward methods 
perform stylization $\sim$1000 times faster than the optimization-based methods. However, they cannot adapt to arbitrary styles that are not used for training. For synthesizing an image from a new style, the entire network needs retraining. To deal with such a restriction, a number of recent approaches encode multiple styles within a single feed-forward network \cite{dumoulin2016,chen2016fast,chen2017stylebank,li2017diversified}.

Dumoulin \emph{et al.} \cite{dumoulin2016} use conditional instance normalization that learns normalization parameters for each style. Given feature activations of the content and style images, \cite{chen2016fast} replaces content features with the closest-matching style features patch-by-patch. Chen \emph{et al.}~\cite{chen2017stylebank} present a network that learns a set of new filters for every new style.   
Li~\emph{et al.}~\cite{li2017diversified} also adapt a single feed-forward network via a texture controller module which forces the network towards synthesizing the desired style only. We note that the existing feed-forward approaches have to compromise between the generalization \cite{li2017diversified,huang2017arbitrary,zhang2017multi} and quality \cite{ulyanov2017improved,ulyanov2016instance,gupta2017characterizing}.

\begin{figure*}[!t]
\centering
\scalebox{1}[1]{\includegraphics[width=0.8\linewidth]{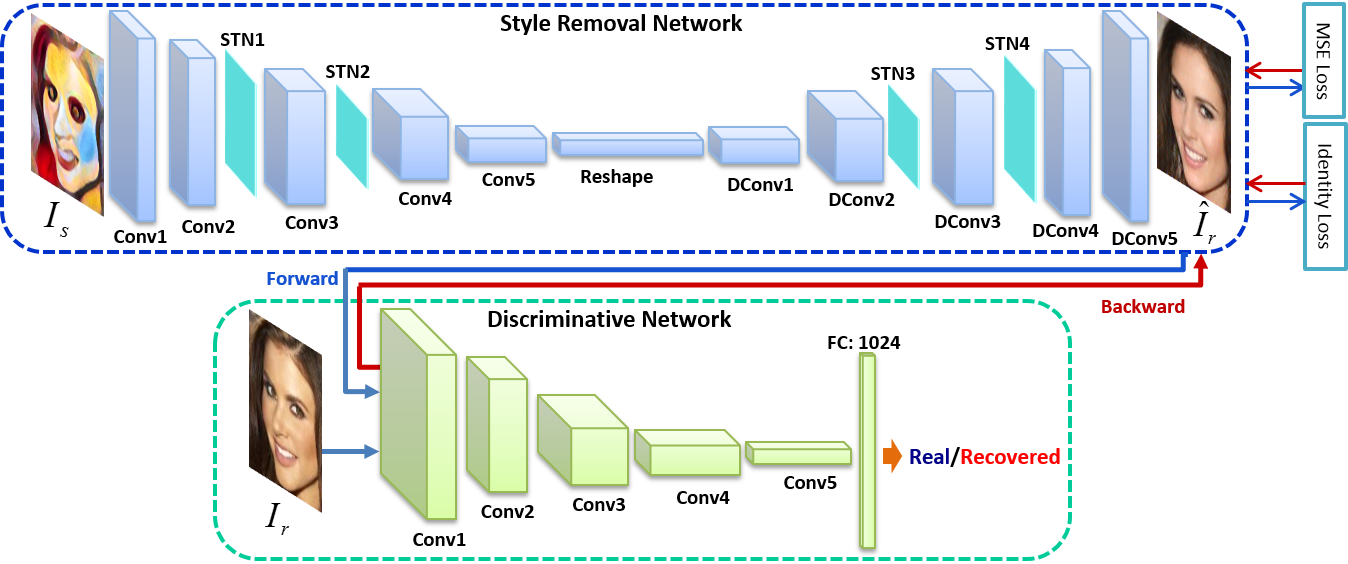}}
\caption{The Architecture of our identity-preserving face destylization framework consists of two parts: a style removal network (blue frame) and a discriminative network (green frame).}
\label{fig:pipeline}
\vspace{-0.4cm}
\end{figure*}

\section{Proposed Method}
We aim to infer a photorealistic and identity-preserving face $\mIh_r$ from an unaligned stylized face $\mI_s$. For this purpose, we design our IFRP framework which contains a Style Removal Network (SRN) and a Discriminative Network (DN). We encourage our SRN to recover faces that come from the latent space of real faces. The DN is trained to distinguish recovered faces from real ones. The general architecture of our IFRP framework is depicted in Figure~\ref{fig:pipeline}.
 
\subsection{Style Removal Network}
Since the goal of face recovery is to generate a photorealistic destylized image, a generative network should be able to remove various styles of portraits without losing the identity-preserving information. To this end, we propose our SRN which comprises an autoencoder (a downsampling encoder and an upsampling decoder) and the STN layers. Figure~\ref{fig:pipeline} shows the architecture of our SRN (enclosed by the blue frame).

The autoencoder learns a deterministic mapping from a portrait space into a latent space with the use of encoder, and a mapping from the latent space to the real face space with the use of decoder.
In this manner, the encoder extracts the high-level features of the unaligned stylized faces and projects them into the feature maps of the real face domain while the decoder synthesizes photorealistic faces from the extracted information.

Considering that the input stylized faces are often misaligned, tilted or rotated \etc, we incorporate four STN layers~\cite{jaderberg2015spatial} to perform face alignments in a data-driven fashion. 
The STN layer can estimate the motion parameters of face images and warp them to a canonical view. The architecture of our STN layers can be found in the supplementary material.
Figure \ref{fig:DiscEffect} illustrates that a successful alignment can be performed by combining STN layers with out network.

\subsection{Discriminative Network}
Using only a pixel-wise distance between the recovered faces and their ground-truth real counterparts leads to over-smoothed results, as shown in Figure~\ref{fig:Disc}. To obtain appealing visual results, we introduce a discriminator, which forces recovered faces to reside in the same latent space as real faces. 
Our proposed DN is composed of convolutional layers and fully connected layers, as illustrated in Figure~\ref{fig:pipeline} (the green frame). 
The discriminative loss, also known as the adversarial loss, penalizes the discrepancy between the distributions of recovered and real faces. This loss is also used to update the parameters of the SRN unit (we alternate over updates of the parameters of SRN and DN). Figure~\ref{fig:Disd} shows the impact of the adversarial loss on the final results.

\subsection{Identity Preservation}
\begin{figure}[t]
\centering
\hspace{-0.5em}
\includegraphics[width=0.18\linewidth]{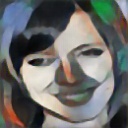}
\includegraphics[width=0.18\linewidth]{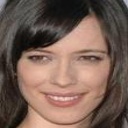}
\includegraphics[width=0.18\linewidth]{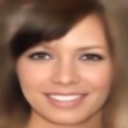}
\includegraphics[width=0.18\linewidth]{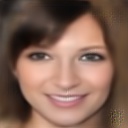}
\includegraphics[width=0.18\linewidth]{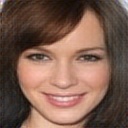}\\
\vspace{-1.5mm}
\hspace{-0.5em}
\subfigure[]{\label{fig:Disa}\scalebox{1}[1]{\includegraphics[width=0.18\linewidth]{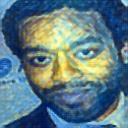}}}
\subfigure[]{\label{fig:Disb}\scalebox{1}[1]
{\includegraphics[width=0.18\linewidth]{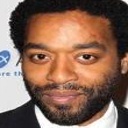}}}
\subfigure[]{\label{fig:Disc}\scalebox{1}[1]
{\includegraphics[width=0.18\linewidth]{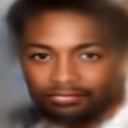}}}
\subfigure[]{\label{fig:Disd}\scalebox{1}[1]
{\includegraphics[width=0.18\linewidth]{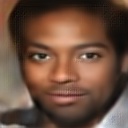}}}
\subfigure[]{\label{fig:Dise}\scalebox{1}[1]
{\includegraphics[width=0.18\linewidth]{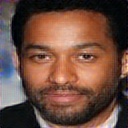}}}
\caption{Contribution of each component of our IFRP network. (a) Input unaligned portraits from unseen styles. (b) Ground-truth face images. (c) Recovered faces with the $\ell_2$ loss. (d) Recovered faces without the identity-preserving loss. (e) Our final results.}
\label{fig:DiscEffect}  
\vspace{-0.4cm}
\end{figure}

By using the adversarial loss, our SRN is able to generate high-frequency facial contents. However, the results often lack details of identities such as the beard or wrinkles, as illustrated in Figure~\ref{fig:Disd}. 
A possible way to address this issue is to constrain the recovered faces to share as many features as possible with the ground-truth faces.

We construct an identity-preserving loss motivated by the ideas of Gatys~\etal~\cite{gatys2016image} and Johnson~\etal~\cite{johnson2016perceptual}. Specifically, we define an Euclidean distance between the feature representations of the recovered and the ground truth image, respectively. The feature maps are obtained from the ReLU activations of the VGG-19 network~\cite{simonyan2014very}. Since the VGG network is pre-trained on a very large image dataset, it can capture visually meaningful facial features. 
Hence, we can preserve the identity information by encouraging the feature similarity between the generated and ground-truth faces. 
We combine the pixel-wise loss, the adversarial loss and the identity-preserving loss together as our final loss function to train our network. Figure~\ref{fig:Dise} illustrates that, with the help of the identity-preserving loss, our IFRP network can reconstruct satisfying identity-preserving results. 
 
\subsection{Training Details}
\label{sec:training}
To train our IFRP network in an end-to-end fashion, we require a large number of SF/RF training image pairs. For each RF, we synthesize different unaligned SF images from various artistic styles to obtain SF/RF $(\mI_s,\mI_r)$ training pairs. As described in Section~\ref{Sec:Data}, we only use stylized faces from three distinct styles in the training stage. 

Our goal is to train a feed-forward network SRN to produce an aligned photorealistic face from any given unaligned portrait. To achieve this, we force the recovered face $\mIh_r$ to be similar to its ground-truth counterpart $\mI_r$. Denote $\mG_\mTheta(\mI_s)$ as the output of our SRN. Since the STN layers are interwoven with the layers of our autoencoder, we optimize the parameters of the autoencoder and the STN layers simultaneously. 
The pixel-wise loss function $\mathcal{L}_{\small{MSE}}$ between $\mIh_r$ and $\mI_r$ is expressed as:
%
\begin{equation}
\label{eqn:MSE}
\begin{split}
\mathcal{L}_{\small{MSE}}(\mTheta)\!&=\!\mathbb{E}_{(\mI_s,\mI_r)\sim p(\mI_s,\mI_r)} \|\mG_{\mTheta}(\mI_s) - \mI_r\|_F^2, \nonumber
\end{split}
\end{equation}
where $p(\mI_s,\mI_r)$ represents the joint distribution of the SF and RF images in the training dataset, and $\mTheta$ denotes the parameters of the SRN unit.

To obtain convincing identity-preserving results, we propose an identity-preserving loss to be the Euclidean distance between the features of recovered face $\mIh_r= \mG_{\mTheta}(\mI_s)$ and ground-truth face $\mI_r$. 
The identity-preserving loss $\mathcal{L}_{id}$ is written as follows: 
\begin{equation}
\label{Identity}
\begin{split}
&\mathcal{L}_{id}(\mTheta)=  \mathbb{E}_{(\mI_s,\mI_r)\sim p(\mI_s,\mI_r)}\|\vpsi(\mG_{\mTheta}(\mI_s))-\vpsi(\mI_r)\|^2_{F}, \nonumber
\end{split}\!\!\!
\end{equation}
where $\vpsi(.)$ denotes the extracted feature maps from the layer ReLU3-2 of the VGG-19 model with respect to some input image.

Motivated by the idea of~\cite{Goodfellow2014,denton2015deep,radford2015unsupervised}, we aim to make the discriminative network $\mD_\mPhi$ fail to distinguish recovered faces from real ones.
Therefore, the parameters of the discriminator $\mPhi$ are updated by minimizing $\mathcal{L}_{dis}$, expressed as:
\begin{equation}
\label{eqn:discr}
\begin{split}
\mathcal{L}_{dis}(\mPhi)\!=&-\!\mathbb{E}_{\mI_r\sim p(\mI_r)}[\log \mD_\mathcal{\mPhi}(\mI_r)]\\
&-\! \mathbb{E}_{\mIh_r\sim p(\mIh_r)}[\log(1\!-\!\mD_\mathcal{\mPhi}(\mIh_r))], \nonumber
\end{split}\!\!\!
\end{equation}
where $p(\mI_r)$ and $p(\mIh_r)$ indicate the distributions of real and recovered faces respectively, and $\mD_\mPhi(\mI_r)$ and $\mD_\mPhi(\mIh_r)$ are the outputs of $\mD_\mPhi$. The $\mathcal{L}_{dis}$ loss is also back-propagated w.r.t. the parameters $\mTheta$ of the SRN unit. 

Our SNR loss is a weighted sum of three terms: the pixel-wise loss, the adversarial loss, and the identity-preserving loss. The parameters $\mTheta$ are obtained by minimizing the objective function of the SRN loss as follows:
\begin{equation}
\begin{split}
\mathcal{L}_{SNR}(\mTheta)=&\!\mathbb{E}_{(\mI_s,\mI_r)\sim p(\mI_s,\mI_r)} \|\mG_{\mTheta}(\mI_s) - \mI_r\|_F^2\\
\!+&\!\lambda\ \mathbb{E}_{\mI_s\sim p(\mI_s))}[\log\!\mD_\mPhi(\mG_{\mTheta}(\mI_s))]\\
\!+&\!\eta\ \mathbb{E}_{(\mI_s,\mI_r)\sim p(\mI_s,\mI_r)}\|\vpsi(\mG_{\mTheta}(\mI_s))-\vpsi(\mI_r)\|^2_{F},
\nonumber
\end{split}
\end{equation}
where $\lambda$ and $\eta$ are trade-off parameters for the discriminator and the identity-preserving losses respectively, and $p(\mI_s)$ is the distribution of stylized faces. 

Since both $\mG_{\mTheta}(\cdot)$ and $\mD_{\mPhi}(\cdot)$ are differentiable functions, the error can be back-propagated w.r.t. $\mTheta$ and $\mPhi$ by the use of the Stochastic Gradient Descent (SGD) combined with Root Mean Square Propagation (RMSprop)~\cite{Hinton}, which helps our algorithm to converge faster. 


\subsection{Implementation Details}
The batch normalization procedure is applied after our convolutional and deconvolutional layers except for the last deconvolutional layer, similar to the models described in ~\cite{Goodfellow2014,radford2015unsupervised}. We also use leaky rectifier with piece-wise linear units (leakyReLU~\cite{maas2013rectifier}) and the negative slope equal $0.2$ as the non-linear activation function. Our network is trained with a mini-batch size of 64. In all our experiments, the parameters $\lambda$ and $\eta$ are set to $10^{-2}$ and $10^{-3}$. We also set the learning rate to $10^{-3}$ and the decay rate to $10^{-2}$. 

As the iterations progress, the images of output faces will be more similar to the ground-truth. Hence, we gradually reduce the effect of the discriminative network by decreasing $\lambda$. Thus,
$\lambda^{n} = \max\{\lambda\cdot 0.995^n, \lambda/2 \}$,
where $n$ is the epoch index.
The strategy of decreasing $\lambda$ not only enriches the effect of the pixel-level similarity but also keeps the discriminative information in the SRN during training.
We also decrease $\eta$ to reduce the impact of the identity-preserving constraint after each iteration: $\eta^{n} = \max\{\eta\cdot 0.995^n, \eta/2\}$.

As our method is feed-forward and no optimization is required at the test time, it takes 10 ms to destylize a 128$\times$128 image. We plan to release the dataset and the code.

\begin{figure}
\begin{minipage}{0.18\linewidth}
\centering
\subfigure[]{\label{fig:dataset1}\scalebox{1}[1]{\includegraphics[width=0.98\linewidth]{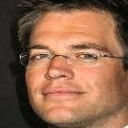}}}
\end{minipage}
\hspace{-0.8 em}
\begin{minipage}{0.75\linewidth}
\centering
\subfigure[]{\label{fig:dataseta}\scalebox{1}[1]{\includegraphics[width=0.23\linewidth]{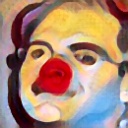}}}
\subfigure[]{\label{fig:datasetb}\scalebox{1}[1]{\includegraphics[width=0.23\linewidth]{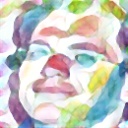}}}
\subfigure[]{\label{fig:datasetc}\scalebox{1}[1]{\includegraphics[width=0.23\linewidth]{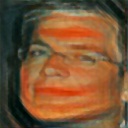}}}
\subfigure[]{\label{fig:datasetd}\scalebox{1}[1]{\includegraphics[width=0.23\linewidth]{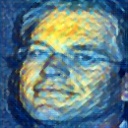}}}\\\vspace{-0.8em}
\subfigure[]{\label{fig:datasete}\scalebox{1}[1]{\includegraphics[width=0.23\linewidth]{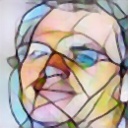}}}
\subfigure[]{\label{fig:datasetf}\scalebox{1}[1]{\includegraphics[width=0.23\linewidth]{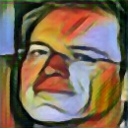}}}
\subfigure[]{\label{fig:datasetg}\scalebox{1}[1]{\includegraphics[width=0.23\linewidth]{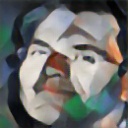}}}
\subfigure[]{\label{fig:dataseth}\scalebox{1}[1]{\includegraphics[width=0.23\linewidth]{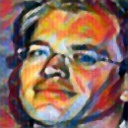}}} \vspace{-0.5em}
\end{minipage}
\caption{Samples of the synthesized dataset. (a) The ground-truth aligned real face image. (b)-(d) The synthesized portraits form~\emph{Candy, Feathers} and \emph{Scream} which have been used for training our network. (e)-(i) The synthesized portraits form \emph{Starry, Mosaic, la Muse, Udnie} and \emph{Composition VII} styles which have not been used for training.}
\label{fig:dataset}
\vspace{-0.4cm}
\end{figure}

\section{Synthesized Dataset and Preprocessing}
\label{Sec:Data}
To train our IFRP network and avoid overfitting, a large number of SF/RF image pairs are required. 
To generate a dataset of such pairs, we employ the CelebA~\cite{Liu2015faceattributes} dataset.
We first randomly choose 10K aligned real faces from the CelebA dataset for training and 1K images for testing. We use these images as our RF ground-truth faces $\mI_r$ which are aligned by eyes. The original size of the images is $178\!\times\!218$ pixels. We crop the central part of each image and resize it to $128\!\times\!128$ pixels. Second, we apply affine transformations to the aligned real faces to generate in-plane unaligned faces.  
To synthesize our training dataset, we retrain the~\textquotedblleft fast style transfer\textquotedblright\ network \cite{johnson2016perceptual} for three different artworks \emph{Scream, Candy} and \emph{Feathers} separately.
Note that recovering photorealistic faces from \emph{Candy, Feathers} and \emph{Scream} styles is more challenging compared to other styles, because facial details are distorted and over-smoothed during the stylization process, as shown in Figure~\ref{fig:dataset}.
Finally, we obtain 30K SF/RF training pairs. 
We also use 1K unaligned real faces to generate 8K SF images from eight diverse styles (\emph{Starry Night, la Muse, Composition VII, Scream, Candy, Feathers, Mosaic} and \emph{Udnie}) as our testing dataset. There is no overlap between the training and testing datasets.




\begin{figure*}[t]
\hspace{4em}
\begin{minipage}{0.085\linewidth}
\centering
\subfigure[RF]{\label{fig:cmp1rf}\scalebox{1}[1]
{\includegraphics[width=1.15\linewidth]{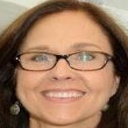}}} \hspace{0.2em}
\end{minipage}
\begin{minipage}{0.85\linewidth}
\centering
\hspace{-3.9em}\includegraphics[width=0.11\linewidth]{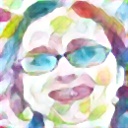}
\includegraphics[width=0.11\linewidth]{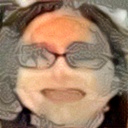}
\includegraphics[width=0.11\linewidth]{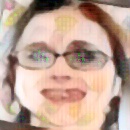}
\includegraphics[width=0.11\linewidth]{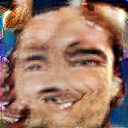}
\includegraphics[width=0.11\linewidth]{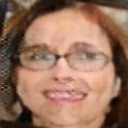}
\includegraphics[width=0.11\linewidth]{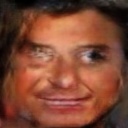}
\includegraphics[width=0.11\linewidth]{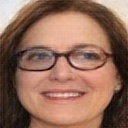}\\\vspace{0.1em}
\hspace{-3.9em}\includegraphics[width=0.11\linewidth]{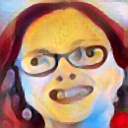}
\includegraphics[width=0.11\linewidth]{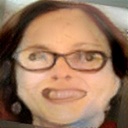}
\includegraphics[width=0.11\linewidth]{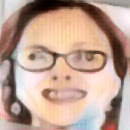}
\includegraphics[width=0.11\linewidth]{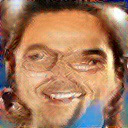}
\includegraphics[width=0.11\linewidth]{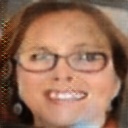}
\includegraphics[width=0.11\linewidth]{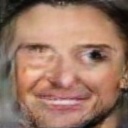}
\includegraphics[width=0.11\linewidth]{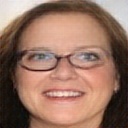}\\\vspace{0.1em}
\hspace{-3.9em}\includegraphics[width=0.11\linewidth]{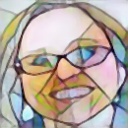}
\includegraphics[width=0.11\linewidth]{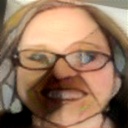}
\includegraphics[width=0.11\linewidth]{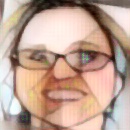}
\includegraphics[width=0.11\linewidth]{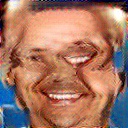}
\includegraphics[width=0.11\linewidth]{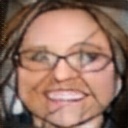}
\includegraphics[width=0.11\linewidth]{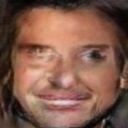}
\includegraphics[width=0.11\linewidth]{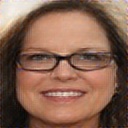}\\\vspace{0.1em}
\hspace{-3.9em}\includegraphics[width=0.11\linewidth]{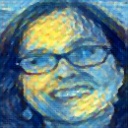}
\includegraphics[width=0.11\linewidth]{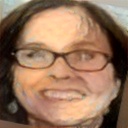}
\includegraphics[width=0.11\linewidth]{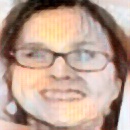}
\includegraphics[width=0.11\linewidth]{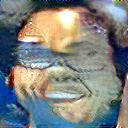}
\includegraphics[width=0.11\linewidth]{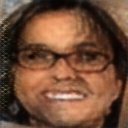}
\includegraphics[width=0.11\linewidth]{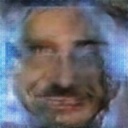}
\includegraphics[width=0.11\linewidth]{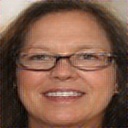}\\\vspace{-1.2mm}
\hspace{-3.9em}\subfigure[SF]{\label{fig:cmp1b}\scalebox{1}[1]{\includegraphics[width=0.11\linewidth]{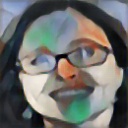}}}
\subfigure[\cite{gatys2016image}]{\label{fig:cmp1c}\scalebox{1}[1]
{\includegraphics[width=0.11\linewidth]{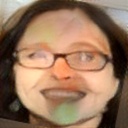}}}
\subfigure[\cite{johnson2016perceptual}]{\label{fig:cmp1d}\scalebox{1}[1]
{\includegraphics[width=0.11\linewidth]{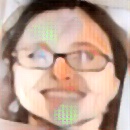}}}
\subfigure[\cite{li2016precomputed}]{\label{fig:cmp1e}\scalebox{1}[1]
{\includegraphics[width=0.11\linewidth]{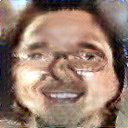}}}
\subfigure[\cite{isola2016image}]{\label{fig:cmp1f}\scalebox{1}[1]
{\includegraphics[width=0.11\linewidth]{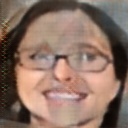}}}
\subfigure[\cite{zhu2017unpaired}]{\label{fig:cmp1g}\scalebox{1}[1]
{\includegraphics[width=0.11\linewidth]{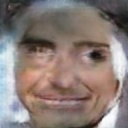}}}
\subfigure[Ours]{\label{fig:cmp1h}\scalebox{1}[1]
{\includegraphics[width=0.11\linewidth]{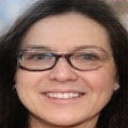}}}
\end{minipage}
\hfill
\vspace{-0.1cm}
\caption{Comparisons of the state-of-the-art methods. (a) The ground-truth real face. (b) Input portraits (from the test dataset) including the seen styles \emph{Feathers} and \emph{Candy} as well as the unseen styles \emph{Mosaic, Starry} and \emph{Udnie}. (c) Gatys~\etal's method~\cite{gatys2016image}. (d) Johnson~\etal's method~\cite{johnson2016perceptual}. (e) Li and Wand's method~\cite{li2016precomputed} (MGAN). (f) Isola~\etal's method~\cite{isola2016image} (pix2pix). (g) Zhu~\etal's method~\cite{zhu2017unpaired} (CycleGAN). (h) Our method.}
\label{fig:cmp1}
\vspace{-0.4cm}
\end{figure*}

\section{Experiments}
\label{expriment}
Below, we compare 
our approach qualitatively and quantitatively to the state-of-the-art methods. To the best of our knowledge, there are no methods which are designed to recover photorealistic faces from portraits. To conduct a fair comparison, we retrain the approaches~\cite{gatys2016image,johnson2016perceptual,li2016precomputed,isola2016image,zhu2017unpaired} on our training dataset for the task of destylization.
\subsection{Qualitative Evaluation}
We visually compare our approach against five  methods detailed below. To let them achieve their best performance, we align SF images in the test dataset (via STN network). 

\begin{figure*}[t]
\hspace{4em}
\begin{minipage}{0.085\linewidth}
\centering
\subfigure[RF]{\label{fig:cmp2rf}\scalebox{1}[1]
{\includegraphics[width=1.15\linewidth]{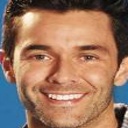}}} \hspace{0.2em}
\end{minipage}
\begin{minipage}{0.85\linewidth}
\centering
\hspace{-3.9em}\includegraphics[width=0.11\linewidth]{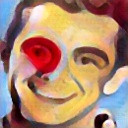} 
\includegraphics[width=0.11\linewidth]{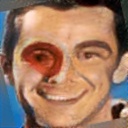} 
\includegraphics[width=0.11\linewidth]{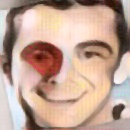} 
\includegraphics[width=0.11\linewidth]{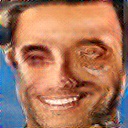}
\includegraphics[width=0.11\linewidth]{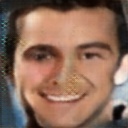} 
\includegraphics[width=0.11\linewidth]{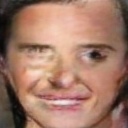} 
\includegraphics[width=0.11\linewidth]{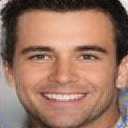}\\\vspace{0.1em}
\hspace{-3.9em}\includegraphics[width=0.11\linewidth]{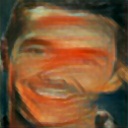} 
\includegraphics[width=0.11\linewidth]{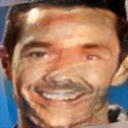} 
\includegraphics[width=0.11\linewidth]{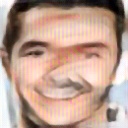}
\includegraphics[width=0.11\linewidth]{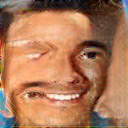}
\includegraphics[width=0.11\linewidth]{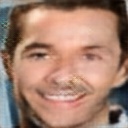}
\includegraphics[width=0.11\linewidth]{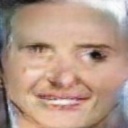}
\includegraphics[width=0.11\linewidth]{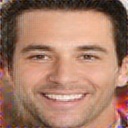}\\\vspace{0.1em}
\hspace{-3.9em}\includegraphics[width=0.11\linewidth]{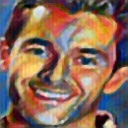} 
\includegraphics[width=0.11\linewidth]{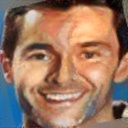} 
\includegraphics[width=0.11\linewidth]{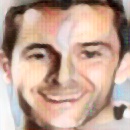} 
\includegraphics[width=0.11\linewidth]{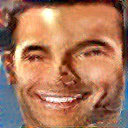} 
\includegraphics[width=0.11\linewidth]{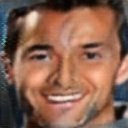} 
\includegraphics[width=0.11\linewidth]{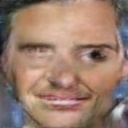} 
\includegraphics[width=0.11\linewidth]{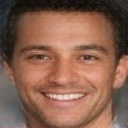}\\\vspace{0.1em}
\hspace{-3.9em}\includegraphics[width=0.11\linewidth]{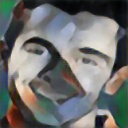} 
\includegraphics[width=0.11\linewidth]{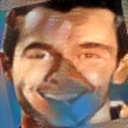} 
\includegraphics[width=0.11\linewidth]{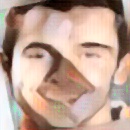} 
\includegraphics[width=0.11\linewidth]{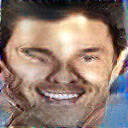} 
\includegraphics[width=0.11\linewidth]{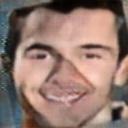} 
\includegraphics[width=0.11\linewidth]{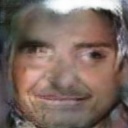} 
\includegraphics[width=0.11\linewidth]{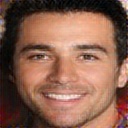}\\\vspace{-1.2mm}
\hspace{-3.9em}\subfigure[SF]{\label{fig:cmp2a}\scalebox{1}[1]{\includegraphics[width=0.11\linewidth]{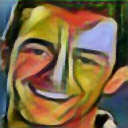}}} 
\subfigure[\cite{gatys2016image}]{\label{fig:cmp2c}\scalebox{1}[1]
{\includegraphics[width=0.11\linewidth]{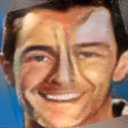}}} 
\subfigure[\cite{johnson2016perceptual}]{\label{fig:cmp2d}\scalebox{1}[1]
{\includegraphics[width=0.11\linewidth]{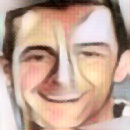}}} 
\subfigure[\cite{li2016precomputed}]{\label{fig:cmp2e}\scalebox{1}[1]
{\includegraphics[width=0.11\linewidth]{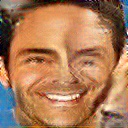}}} 
\subfigure[\cite{isola2016image}]{\label{fig:cmp2f}\scalebox{1}[1]
{\includegraphics[width=0.11\linewidth]{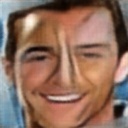}}} 
\subfigure[\cite{zhu2017unpaired}]{\label{fig:cmp2g}\scalebox{1}[1]
{\includegraphics[width=0.11\linewidth]{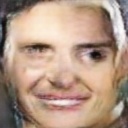}}} 
\subfigure[Ours]{\label{fig:cmp2h}\scalebox{1}[1]
{\includegraphics[width=0.11\linewidth]{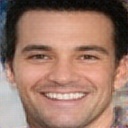}}} 
\end{minipage}
\hfill
\vspace{-0.1cm}
\caption{(a) The ground-truth real face. (b) Input portraits (from the test dataset) including the seen styles \emph{Candy} and\emph{Scream} as well as the unseen styles \emph{Composition VII, Udnie} and \emph{la Muse} from unseen styles. (c) Gatys~\etal's method~\cite{gatys2016image}. (d) Johnson~\etal's method~\cite{johnson2016perceptual}. (e) Li and Wand's method~\cite{li2016precomputed} (MGAN). (f) Isola~\etal's method~\cite{isola2016image} (pix2pix). (g) Zhu~\etal's method~\cite{zhu2017unpaired} (CycleGAN). (h) Our method.}
\label{fig:cmp2}
\vspace{-0.3cm}
\end{figure*}

Gatys \emph{et al.}~\cite{gatys2016image} is an image-optimization based style transfer method which does not have any training stage.
This method captures the correlation between feature maps of the portrait and the synthesized face (Gram matrices) in different layers of a CNN. Therefore, spatial structures of face images cannot be preserved. 
As shown in Figures~\ref{fig:cmp1c} and \ref{fig:cmp2c}, the network fails to produce realistic results and the artistic styles have not been fully removed.

We retrain the approach proposed by Johnson \emph{et al.} \cite{johnson2016perceptual} for destylization. Due to the use of the Gram matrix, their network also generates distorted facial details and produces unnatural effects. As shown in Figures~\ref{fig:cmp1d} and \ref{fig:cmp2d}, the facial details are blurred and the skin colors are not homogeneous. As shown in the first row of Figure~\ref{fig:cmp2d}, we observe that the styles of the eyes were not removed from outputs.  

MGAN~\cite{li2016precomputed} is a patch-based style transfer method. We retrain this network for the purpose of the face recovery. As this method is trained on RF/SF patches, it cannot capture the global structure of entire faces. As seen in Figures~\ref{fig:cmp1e} and \ref{fig:cmp2e}, this method produces distorted results and the facial colors are inconsistent. In contrast, our method successfully captures the global structure of faces and generates highly-consistent facial colors.

Isola~\emph{et al.}~\cite{isola2016image} train a "U-net" generator augmented with a PatchGAN discriminator in an adversarial framework, known as "pix2pix". Since the patch-based discriminator is trained to classify whether an image patch is sampled from real faces or not, this network does not take the global structure of faces into account.
In addition, the U-net concatenates low-level features from the bottom layers of the encoder with the features in the decoder to generate face images. Because the low-level features of input images are passed to the outputs, this network fails to eliminate the artistic styles in the face images.
As shown in Figures~\ref{fig:cmp1f} and \ref{fig:cmp2f}, although pix2pix can generate acceptable results for the seen styles, it fails to remove the unseen styles and produces obvious artifacts . 

CycleGAN~\cite{zhu2017unpaired} is an image-to-image translation method that uses unpaired datasets. This network provides a mapping between two different domains by the use of a cycle-consistency loss.
Since CycleGAN also employs a patch-based discriminator, this network cannot capture the global structure of faces. As this network uses unpaired face datasets \ie, unpaired RF and SF images, the low-level features of the stylized faces and real faces are uncorrelated.
Thus, CycleGAN is not suitable for transferring stylized portraits to photorealistic ones. As shown in Figures~\ref{fig:cmp1g} and ~\ref{fig:cmp2g}, this method produces distorted results and does not preserve the identities with respect to the input images.

In contrast, our results demonstrate higher fidelity and better consistency with respect to the real faces, such as facial expressions and skin colors. Our network can preserve identity information of a subject for both seen and unseen styles, as shown in Figures~\ref{fig:cmp1h} and ~\ref{fig:cmp2h}.

\subsection{Quantitative Evaluation}

\noindent{\textbf{Pixel-wise Recovery Analysis:}}

To evaluate the pixel-wise recovery performance, we use the average Peak Signal to Noise Ratio (PSNR) and Structural Similarity (SSIM)~\cite{wang2004image} scores on seen and unseen styles of our test dataset. The pixel-wise recovery results for each method are summarized in Table~\ref{tab1} (higher scores indicate better results). 
The PSNR and SSIM scores confirm that our IFRP approach outperforms other state-of-the-art methods on both seen (the first and second rows) and unseen (the third, fourth and fifth rows) styles. Figures~\ref{fig:cmp1} and \ref{fig:cmp2} verify the performance visually.
Moreover, we also apply different methods on sketches from the CUFSF dataset as an unseen style without fine-tuning or re-training our network.

In order to demonstrate the contributions of each loss function to the quantitative results, we also show the results for when only the $\ell_2$ loss is used, as indicated by SRN in Table~\ref{tab1}, and for both the $\ell_2$ and discriminative losses, as indicated by SRN+DN in Table~\ref{tab1}. The $\ell_2$ loss considers the intensity similarity only, thus it produces over-smooth faces. The discriminative loss further forces the generated faces to be realistic, thus it improves the final results qualitatively and quantitatively. Benefiting from our combined loss, our network not only achieves highest quantitative results but also generates photorealistic face images.


\begin{table}\renewcommand{\arraystretch}{1}
\centering
\caption{Comparisons of PSNR and SSIM on the entire test dataset.}
\scalebox{0.8}{
\begin{tabular}{c|c|c|c|c|c|c}
\hline
\multirow{2}{*}{Method} & \multicolumn{2}{c|}{Seen Styles} & \multicolumn{2}{c|}{Unseen Styles} & \multicolumn{2}{c}{Unseen Sketches} \\
\cline{2-7}
& PSNR & SSIM & PSNR & SSIM  & PSNR & SSIM \\
\hline
Gatys ~\cite{gatys2016image} & 23.88 & 0.84 & 23.25 & 0.83 & 23.33 & 0.82\\ 
Johnson~\cite{johnson2016perceptual} & 19.65 & 0.82 & 19.81 & 0.81 & 19.77 & 0.82 \\
MGAN~\cite{li2016precomputed} & 20.87 & 0.79 & 20.21 & 0.66 & 21.01 & 0.71 \\
pix2pix~\cite{isola2016image} & 25.28 & 0.89 & 23.10 & 0.85 & 23.88 & 0.86 \\
CycleGAN~\cite{zhu2017unpaired} & 19.584 & 0.78 & 18.99 & 0.77 & 19.60 & 0.77\\
\hline
SRN & 25.12 & 0.89 & 24.09 & 0.88 & 24.13 & 0.89 \\
SRN + DN & 25.25 & 0.90 & 24.25 & 0.89 & 24.56 & 0.90 \\
\bf{IFRP} & {\bf 27.08} & {\bf 0.93} & {\bf 24.83} & {\bf 0.91} & {\bf 24.89} & {\bf 0.92}\\
\hline
\end{tabular}}
\label{tab1}
\vspace{-0.4cm}
\end{table}

\vspace{0.05cm}
\noindent\textbf{Face Retrieval Analysis:}

In this section, we demonstrate that the faces recovered by our method are highly consistent with their ground-truth counterparts.
To this end, we run a face recognition algorithm~\cite{parkhi2015deep} on our test dataset for both seen and unseen styles. For each investigated method, we set 1K recovered faces from one style as a query dataset and then set 1K of ground-truth faces as a search dataset. 
We apply \cite{parkhi2015deep} to quantify whether the correct person is retrieved within the top-5 matched images. Then an average retrieval score is obtained. We repeat this procedure for every style and then obtain the average Face Retrieval Ratio (FRR) by averaging all scores from the seen and unseen styles, respectively. 
As indicated in Table~\ref{tab2}, our IFRP network outperforms the other methods across all the styles. Even for the unseen styles, our method can still retain most identity features, making the destylized results similar to the ground-truth faces.
Moreover, we also run an experiment on hand-drawn sketches of the CUFSF dataset used as an unseen style. The FRR scores are better compared to results on other styles as facial components are easier to extract from sketches/their contours. 
Despite our method is not dedicated to face retrieval, we compare it to \cite{zhang2011coupled}. To challenge our method, we did not re-train our network on sketches (we used other styles). Thus, we recovered faces from sketches (CUFSF dataset) and performed face identification that yielded $\sim$91\% Verification Rate (VR) \@ FAR=0.1\%. This outperforms photo-synthesizing method MRF+LE \cite{zhang2011coupled} (43.66$\%$ VR at FAR=0.1$\%$) which uses sketches for training.

\vspace{0.05cm}
\noindent{\textbf{Consistency Analysis w.r.t. Styles:}}

As shown in Figures~\ref{fig:cmp1h} and \ref{fig:cmp2h}, our network recovers the photorealistic faces from various stylized portraits of the same person. Note that recovered faces resemble each other. It indicates that our network is robust to different styles. 

In order to demonstrate the robustness of our network to different styles quantitatively, we study the consistency of faces recovered from different styles.
Here, we choose 1K faces destylized from one style. For each destylized face we search its top-5 most similar faces in another group of destylized faces. If the same person is retrieved within the top-5 candidates, we record it as a hit. Then an average hit number of one style is obtained. We repeat the same procedure for all the other 7 styles, and then calculate the average hit number, denoted as Face Consistency Ratio (FCR). Note that the probability of one hit by chance is 0.5$\%$.
Table~\ref{tab2} shows the average FCR scores on the test dataset for each method. The FCR scores indicate that our IFRP method produces the most consistent destylized faces across different styles. This also implies that our SRN can extract facial features irrespective of image styles.

\begin{table}\renewcommand{\arraystretch}{1}
\vspace{-0.1cm}
\centering
\caption{Comparisons of FRR and FCR on the entire test dataset.}
\scalebox{0.8}{
\begin{tabular}{>{\centering\arraybackslash}m{0.245\linewidth}|>{\centering\arraybackslash}m{0.20\linewidth}|>{\centering\arraybackslash}m{0.24\linewidth}|>{\centering\arraybackslash}m{0.26\linewidth}|>{\centering\arraybackslash}m{0.09\linewidth}}
\hline
\multirow{2}{*}{Method} & \multicolumn{3}{c|}{FRR} & \multirow{2}{*}{FCR} \\
\cline{2-4}
& Seen Styles & Unseen Styles & Unseen Sketch & \\
\hline
Gatys ~\cite{gatys2016image}  & 64.67\%  & 60.28\% & 68.36\% & 72.89\%\\ 
Johnson~\cite{johnson2016perceptual}  & 50.54\% &  38.87\% & 40.27\% & 44.99\% \\
MGAN~\cite{li2016precomputed}  & 6.97\% &   12.52\% & 17.99\% & 38.24\% \\
pix2pix~\cite{isola2016image} & 75.13\%  & 59.98\% & 61.63\% & 87.73\%  \\
CycleGAN~\cite{zhu2017unpaired} &  1.07\% &  0.68\% & 0.70\% & 13.32\% \\
\hline
IFRP & {\bf 86.93\%} &  {\bf 74.52\%} & {\bf 91.05\%} & {\bf 92.06\%} \\
\hline
\end{tabular}}
\label{tab2}
\vspace{-0.1cm}
\end{table}

\begin{figure}
\centering
\includegraphics[width=0.16\linewidth]{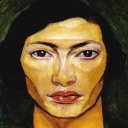}
\includegraphics[width=0.16\linewidth]{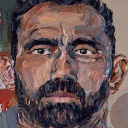}
\includegraphics[width=0.16\linewidth]{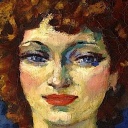}
\includegraphics[width=0.16\linewidth]{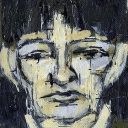}
\includegraphics[width=0.16\linewidth]{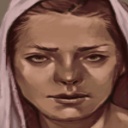}\\\vspace{0.2em}
\includegraphics[width=0.16\linewidth]{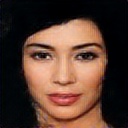}
\includegraphics[width=0.16\linewidth]{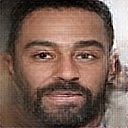}
\includegraphics[width=0.16\linewidth]{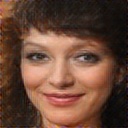}
\includegraphics[width=0.16\linewidth]{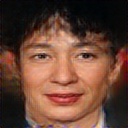}
\includegraphics[width=0.16\linewidth]{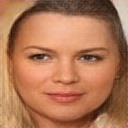}
\caption{Results for the original unaligned paintings. Top row: the original portraits from art galleries. Bottom row: our results.}
\label{fig:Orig}
\vspace{-0.1cm}
\end{figure}

\begin{figure}[t]
\centering
\includegraphics[width=0.16\linewidth]{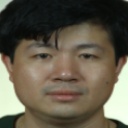}
\includegraphics[width=0.16\linewidth]{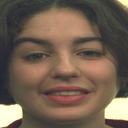}
\includegraphics[width=0.16\linewidth]{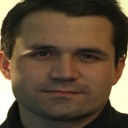}
\includegraphics[width=0.16\linewidth]{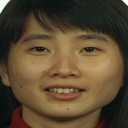}\\
\includegraphics[width=0.16\linewidth]{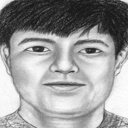}
\includegraphics[width=0.16\linewidth]{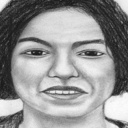}
\includegraphics[width=0.16\linewidth]{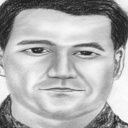}
\includegraphics[width=0.16\linewidth]{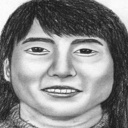}\\
\includegraphics[width=0.16\linewidth]{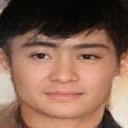}
\includegraphics[width=0.16\linewidth]{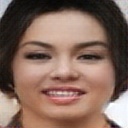}
\includegraphics[width=0.16\linewidth]{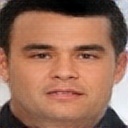}
\includegraphics[width=0.16\linewidth]{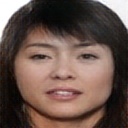}\\
\caption{Recovering photo-realistic faces from hand-drawn sketches from the FERET dataset. Top row: ground-truth faces. Middle row: sketches. Bottom row: our results.}
\label{fig:sketch}
\vspace{-0.4cm}
\end{figure}

\subsection{Destylizing Original Paintings and Sketches}
We demonstrate that our method is not restricted to recovery of faces from computer-generated stylized portraits but it can also deal with real paintings and sketches. To confirm this, we randomly choose a few of paintings from art galleries such as Archibald~\cite{archibald} and hand-drawn sketches from FERET dataset~\cite{phillips1998feret}. Next, we crop face regions from them as our real test images. 
Figures \ref{fig:Orig} and \ref{fig:sketch} show that our method can efficiently recover photorealistic faces. This indicates that our method is not limited to the synthesized data and does not require an alignment procedure beforehand.

\subsection{Limitations} 
\begin{figure}[t]
\centering
\includegraphics[width=0.16\linewidth]{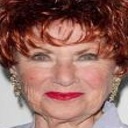}
\includegraphics[width=0.16\linewidth]{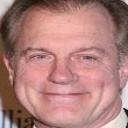}
\includegraphics[width=0.16\linewidth]{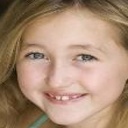}
\includegraphics[width=0.16\linewidth]{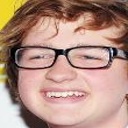}\\
\includegraphics[width=0.16\linewidth]{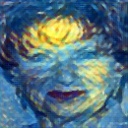}
\includegraphics[width=0.16\linewidth]{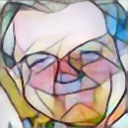}
\includegraphics[width=0.16\linewidth]{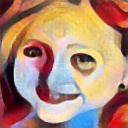}
\includegraphics[width=0.16\linewidth]{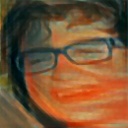}\\
\includegraphics[width=0.16\linewidth]{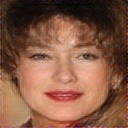}
\includegraphics[width=0.16\linewidth]{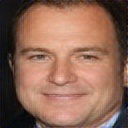}
\includegraphics[width=0.16\linewidth]{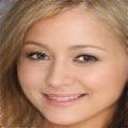}
\includegraphics[width=0.16\linewidth]{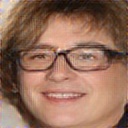}\\
\caption{Limitations. Top row: ground-truth faces. Middle row: unaligned stylized faces. Bottom row: our results.}
\label{fig:Failure}
\vspace{-0.5cm}
\end{figure}

We note that in the CelebA dataset, numbers of images of children, old people and young adults are unbalanced \eg, there are more images of young adults than children and old people. This makes our synthesized dataset unbalanced. Hence, facial features of children and old people is are not fully represented in our dataset. Therefore, our network may be prone to recover images with facial features of young adults for children and old people, as seen in Figure~\ref{fig:Failure}. In addition, because the color information has been distorted in the stylized paintings, it is very challenging to recover the skin and hair color that is consistent with the ground-truth without introducing additional cues.
In future, we intend to embed semantic information into our network and then generate more consistent face images in terms of the skin and hair color.

\section{Conclusion}

We introduce a novel neural network for face recovery. It extracts features from a given unaligned stylized portrait and then recovers a photorealistic face from these features. The SRN successfully learns a mapping from unaligned stylized faces to aligned photorealistic faces. Moreover, our identity-preserving loss further encourages our network to generate identity trustworthy faces. This makes our algorithm readily available for tasks such as face recognition. We also show that our approach can recover latent faces of portraits in unseen styles, real paintings and sketches. 

\section*{Acknowledgement}
This work is supported by the Australian Research Council (ARC) grant DP150104645.


{\small
\bibliographystyle{ieee}
\bibliography{egbib}
}
\nopagebreak
\appendix

\section{\textbf{Supplementary Material}}

\noindent{\textbf{Face Alignment: Spatial Transfer Networks (STN)}}

As described in Section {\color{red}{3.1}} of the main paper, we incorporate multiple STNs~\cite{jaderberg2015spatial} as intermediate layers to compensate for misalignments and in-plane rotations.  
The STN layers can estimate the motion parameters of face images and warp them to a canonical view. 
STN contains localization, grid generator and sampler modules. The localization module consists of several hidden layers to estimates the transformation parameters with respect to the canonical view. The grid generator module creates a sampling grid according to the estimated parameters. Finally, the sampler module maps the input feature maps into generated girds using the bilinear interpolation.
The architectures of our STN layers are detailed in Tables~\ref{tab3},~\ref{tab4},~\ref{tab5} and ~\ref{tab6}.\\

\begin{table}[!ht]\renewcommand{\arraystretch}{1.2} 
\caption{The STN1 architecture}
\begin{center}
\begin{tabular}{>{\centering\arraybackslash}p{0.9\linewidth}}
\hline
STN1 \\
\hline
 Input: 64 x 64 x 32  \\
3 x 3 x 64 conv, relu, Max-pooling(2,2)  \\
3 x 3 x 128 conv, relu, Max-pooling(2,2) \\
3 x 3 x 256 conv, relu, Max-pooling(2,2)  \\
3 x 3 x 20 conv, relu, Max-pooling(2,2)  \\
3 x 3 x 20 conv, relu\\
fully connected (80,20), relu \\
fully connected (20,4) \\
\hline
\end{tabular}
\label{tab3}
\end{center}
\vspace{-1.1em}
\end{table}
\begin{table}[!ht]\renewcommand{\arraystretch}{1.2}
\caption{The STN2 architecture}
\begin{center}
\begin{tabular}{>{\centering\arraybackslash}p{0.9\linewidth}}
\hline
STN2 \\
\hline
Input: 32 x 32 x 64  \\
3 x 3 x 128 conv, relu, Max-pooling(2,2) \\
3 x 3 x 256 conv, relu, Max-pooling(2,2)  \\
3 x 3 x 20 conv, relu, Max-pooling(2,2)  \\
3 x 3 x 20 conv, relu\\
fully connected (80,20), relu  \\
fully connected (20,4) \\
\hline
\end{tabular}
\label{tab4}
\end{center}
\vspace{-1.1em}
\end{table}
\begin{table}[!ht]\renewcommand{\arraystretch}{1.2}
\caption{The STN3 architecture}
\begin{center}
\begin{tabular}{>{\centering\arraybackslash}p{0.9\linewidth}}
\hline
STN3 \\
\hline
Input: 16 x 16 x 128 \\
3 x 3 x 256 conv, relu, Max-pooling(2,2)  \\
3 x 3 x 20 conv, relu, Max-pooling(2,2)  \\
3 x 3 x 20 conv, relu\\
fully connected (80,20), relu  \\
fully connected (20,4)\\
\hline
\end{tabular}
\label{tab5}
\end{center}
\vspace{-1.1em}
\end{table}
\begin{table}[!ht]\renewcommand{\arraystretch}{1.2}
\caption{The STN4 architecture}
\begin{center}
\begin{tabular}{>{\centering\arraybackslash}p{0.9\linewidth}}
\hline
 STN4\\
\hline
Input: 32 x 32 x 64\\
3 x 3 x 64 conv, relu, Max-pooling(2,2)\\
3 x 3 x 128 conv, relu, Max-pooling(2,2)  \\
3 x 3 x 256 conv, relu, Max-pooling(2,2)  \\
3 x 3 x 20 conv, relu\\
fully connected (80,20), relu \\
fully connected (20,4)\\
\hline
\end{tabular}
\label{tab6}
\end{center}
\vspace{-1.1em}
\end{table}


\noindent{\textbf{Contribution of each component in the IFRP network}}\\

\begin{figure*}[!ht]
\centering
\hspace{-0.5em}\includegraphics[width=0.13\linewidth]{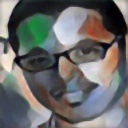}
\includegraphics[width=0.13\linewidth]{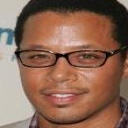}
\includegraphics[width=0.13\linewidth]{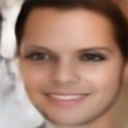}
\includegraphics[width=0.13\linewidth]{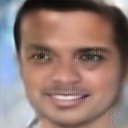}
\includegraphics[width=0.13\linewidth]{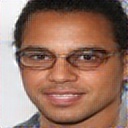}\\
\hspace{-0.5em}\includegraphics[width=0.13\linewidth]{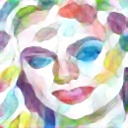}
\includegraphics[width=0.13\linewidth]{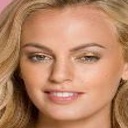}
\includegraphics[width=0.13\linewidth]{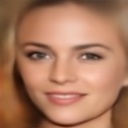}
\includegraphics[width=0.13\linewidth]{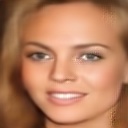}
\includegraphics[width=0.13\linewidth]{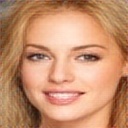}\\
\hspace{-0.5em}\includegraphics[width=0.13\linewidth]{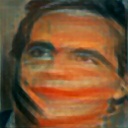}
\includegraphics[width=0.13\linewidth]{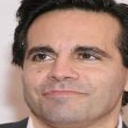}
\includegraphics[width=0.13\linewidth]{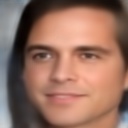}
\includegraphics[width=0.13\linewidth]{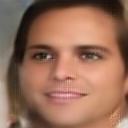}
\includegraphics[width=0.13\linewidth]{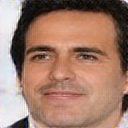}\\
\vspace{-1.5mm}\hspace{-0.5em}
\subfigure[]{\label{fig:Dis1a}\scalebox{1}[1]{\includegraphics[width=0.13\linewidth]{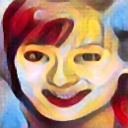}}}
\subfigure[]{\label{fig:Dis1b}\scalebox{1}[1]
{\includegraphics[width=0.13\linewidth]{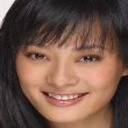}}}
\subfigure[]{\label{fig:Dis1c}\scalebox{1}[1]
{\includegraphics[width=0.13\linewidth]{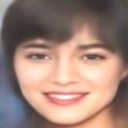}}}
\subfigure[]{\label{fig:Dis1d}\scalebox{1}[1]
{\includegraphics[width=0.13\linewidth]{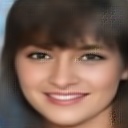}}}
\subfigure[]{\label{fig:Dis1f}\scalebox{1}[1]
{\includegraphics[width=0.13\linewidth]{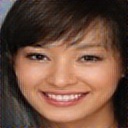}}}
\caption{More results showing contribution of each component in the IFRP network. (a) Input portraits from \emph{Udnie, Feathers, Scream} and \emph{Candy} styles. (b) Ground-truth real faces. (c) Faces recovered by the use of $\ell_2$ loss. (d) Faces recovered by the use of the $\ell_2$ and the adversarial losses. (e) Our final results with the $\ell_2$, the adversarial and the identity-preserving losses.}
\label{fig:contribution}  
\end{figure*}

In Section {\color{red}3} of the main paper, we described impact of the $\ell_2$ loss, the adversarial loss and the identity-preserving loss on the face recovery from portraits. 
Figure \ref{fig:contribution} further shows the contribution of each loss function in the final results.\\

\begin{figure*}[!ht]
\hspace{4em}
\begin{minipage}{0.089\linewidth}
\centering
\subfigure[RF]{\label{fig:cmp3rf}\scalebox{1}[1]
{\includegraphics[width=1.15\linewidth]{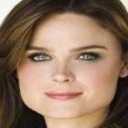}}} \hspace{0.2em}
\end{minipage}
\begin{minipage}{0.89\linewidth}
\centering
\hspace{-3.9em}\includegraphics[width=0.11\linewidth]{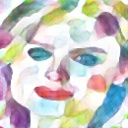}
\includegraphics[width=0.11\linewidth]{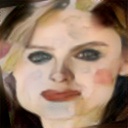}
\includegraphics[width=0.11\linewidth]{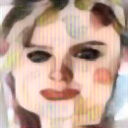}
\includegraphics[width=0.11\linewidth]{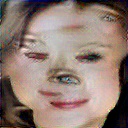}
\includegraphics[width=0.11\linewidth]{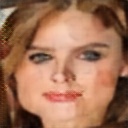}
\includegraphics[width=0.11\linewidth]{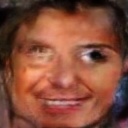}
\includegraphics[width=0.11\linewidth]{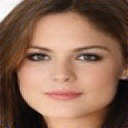}\\\vspace{0.1em}
\hspace{-3.9em}\includegraphics[width=0.11\linewidth]{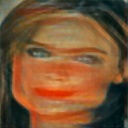}
\includegraphics[width=0.11\linewidth]{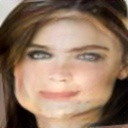}
\includegraphics[width=0.11\linewidth]{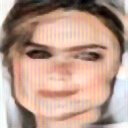}
\includegraphics[width=0.11\linewidth]{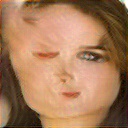}
\includegraphics[width=0.11\linewidth]{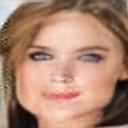}
\includegraphics[width=0.11\linewidth]{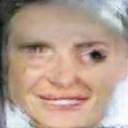}
\includegraphics[width=0.11\linewidth]{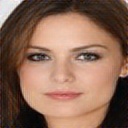}\\\vspace{0.1em}
\hspace{-3.9em}\includegraphics[width=0.11\linewidth]{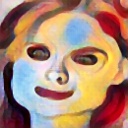}
\includegraphics[width=0.11\linewidth]{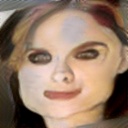}
\includegraphics[width=0.11\linewidth]{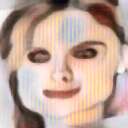}
\includegraphics[width=0.11\linewidth]{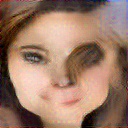}
\includegraphics[width=0.11\linewidth]{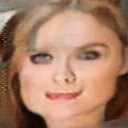}
\includegraphics[width=0.11\linewidth]{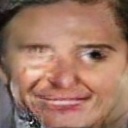}
\includegraphics[width=0.11\linewidth]{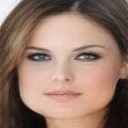}\\\vspace{0.1em}
\hspace{-3.9em}\includegraphics[width=0.11\linewidth]{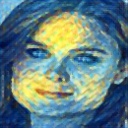}
\includegraphics[width=0.11\linewidth]{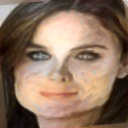}
\includegraphics[width=0.11\linewidth]{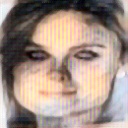}
\includegraphics[width=0.11\linewidth]{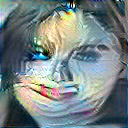}
\includegraphics[width=0.11\linewidth]{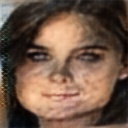}
\includegraphics[width=0.11\linewidth]{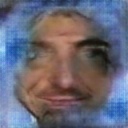}
\includegraphics[width=0.11\linewidth]{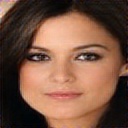}\\\vspace{0.1em}
\hspace{-3.9em}\includegraphics[width=0.11\linewidth]{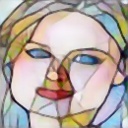}
\includegraphics[width=0.11\linewidth]{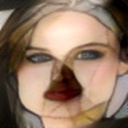}
\includegraphics[width=0.11\linewidth]{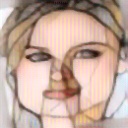}
\includegraphics[width=0.11\linewidth]{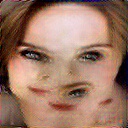}
\includegraphics[width=0.11\linewidth]{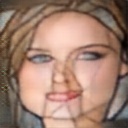}
\includegraphics[width=0.11\linewidth]{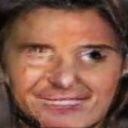}
\includegraphics[width=0.11\linewidth]{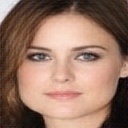}\\\vspace{0.1em}
\hspace{-3.9em}\includegraphics[width=0.11\linewidth]{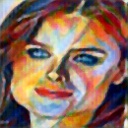}
\includegraphics[width=0.11\linewidth]{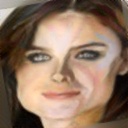}
\includegraphics[width=0.11\linewidth]{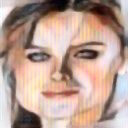}
\includegraphics[width=0.11\linewidth]{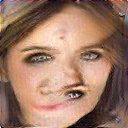}
\includegraphics[width=0.11\linewidth]{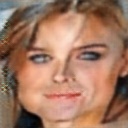}
\includegraphics[width=0.11\linewidth]{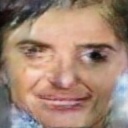}
\includegraphics[width=0.11\linewidth]{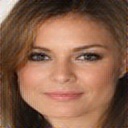}\\\vspace{0.1em}
\hspace{-3.9em}\includegraphics[width=0.11\linewidth]{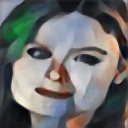}
\includegraphics[width=0.11\linewidth]{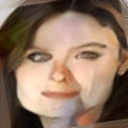}
\includegraphics[width=0.11\linewidth]{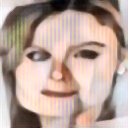}
\includegraphics[width=0.11\linewidth]{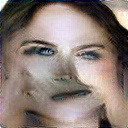}
\includegraphics[width=0.11\linewidth]{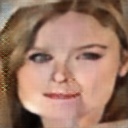}
\includegraphics[width=0.11\linewidth]{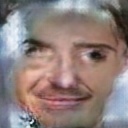}
\includegraphics[width=0.11\linewidth]{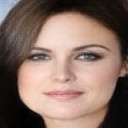}\\\vspace{-1.2mm}
\hspace{-3.9em}\subfigure[SF]{\label{fig:cmp3b}\scalebox{1}[1]{\includegraphics[width=0.11\linewidth]{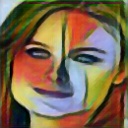}}}
\subfigure[\cite{gatys2016image}]{\label{fig:cmp3c}\scalebox{1}[1]
{\includegraphics[width=0.11\linewidth]{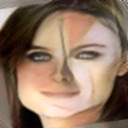}}}
\subfigure[\cite{johnson2016perceptual}]{\label{fig:cmp3d}\scalebox{1}[1]
{\includegraphics[width=0.11\linewidth]{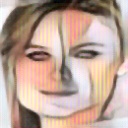}}}
\subfigure[\cite{li2016precomputed}]{\label{fig:cmp3e}\scalebox{1}[1]
{\includegraphics[width=0.11\linewidth]{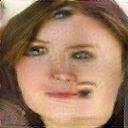}}}
\subfigure[\cite{isola2016image}]{\label{fig:cmp3f}\scalebox{1}[1]
{\includegraphics[width=0.11\linewidth]{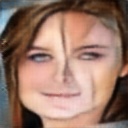}}}
\subfigure[\cite{zhu2017unpaired}]{\label{fig:cmp3g}\scalebox{1}[1]
{\includegraphics[width=0.11\linewidth]{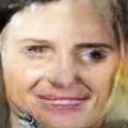}}}
\subfigure[Ours]{\label{fig:cmp3h}\scalebox{1}[1]
{\includegraphics[width=0.11\linewidth]{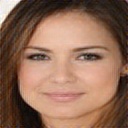}}}
\end{minipage}
\hfill
\vspace{-0.1cm}
\caption{Qualitative comparisons of the state-of-the-art methods. (a) The ground-truth real face. (b) Input portraits (from the test dataset) including the seen styles \emph{Feathers, Scream} and \emph{Candy} as well as the unseen styles \emph{Starry, Mosaic, Composition VII, Udnie} and \emph{La Muse}. (c) Gatys~\etal's method~\cite{gatys2016image}. (d) Johnson~\etal's method~\cite{johnson2016perceptual}. (e) Li and Wand's method~\cite{li2016precomputed} (MGAN). (f) Isola~\etal's method~\cite{isola2016image} (pix2pix). (g) Zhu~\etal's method~\cite{zhu2017unpaired} (CycleGAN). (h) Our method.}
\label{fig:cmp3}
\vspace{-0.3cm}
\end{figure*}

\begin{figure*}[!ht]
\hspace{4em}
\begin{minipage}{0.089\linewidth}
\centering
\subfigure[RF]{\label{fig:cmp4rf}\scalebox{1}[1]
{\includegraphics[width=1.15\linewidth]{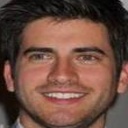}}} \hspace{0.2em}
\end{minipage}
\begin{minipage}{0.89\linewidth}
\centering
\hspace{-3.9em}\includegraphics[width=0.11\linewidth]{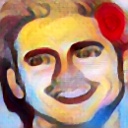}
\includegraphics[width=0.11\linewidth]{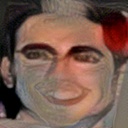}
\includegraphics[width=0.11\linewidth]{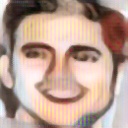}
\includegraphics[width=0.11\linewidth]{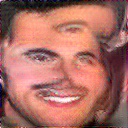}
\includegraphics[width=0.11\linewidth]{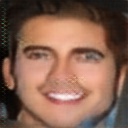}
\includegraphics[width=0.11\linewidth]{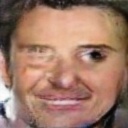}
\includegraphics[width=0.11\linewidth]{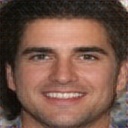}\\\vspace{0.1em}
\hspace{-3.9em}\includegraphics[width=0.11\linewidth]{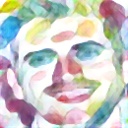}
\includegraphics[width=0.11\linewidth]{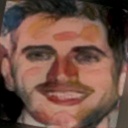}
\includegraphics[width=0.11\linewidth]{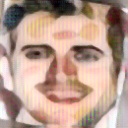}
\includegraphics[width=0.11\linewidth]{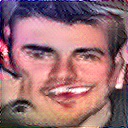}
\includegraphics[width=0.11\linewidth]{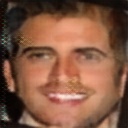}
\includegraphics[width=0.11\linewidth]{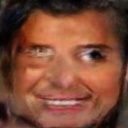}
\includegraphics[width=0.11\linewidth]{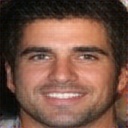}\\\vspace{0.1em}
\hspace{-3.9em}\includegraphics[width=0.11\linewidth]{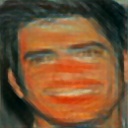}
\includegraphics[width=0.11\linewidth]{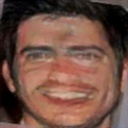}
\includegraphics[width=0.11\linewidth]{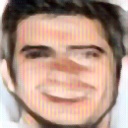}
\includegraphics[width=0.11\linewidth]{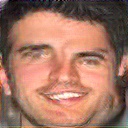}
\includegraphics[width=0.11\linewidth]{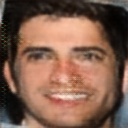}
\includegraphics[width=0.11\linewidth]{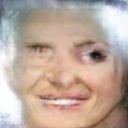}
\includegraphics[width=0.11\linewidth]{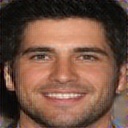}\\\vspace{0.1em}
\hspace{-3.9em}\includegraphics[width=0.11\linewidth]{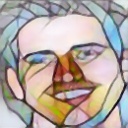}
\includegraphics[width=0.11\linewidth]{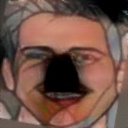}
\includegraphics[width=0.11\linewidth]{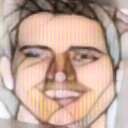}
\includegraphics[width=0.11\linewidth]{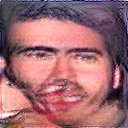}
\includegraphics[width=0.11\linewidth]{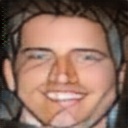}
\includegraphics[width=0.11\linewidth]{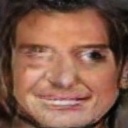}
\includegraphics[width=0.11\linewidth]{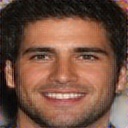}\\\vspace{0.1em}
\hspace{-3.9em}\includegraphics[width=0.11\linewidth]{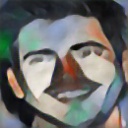}
\includegraphics[width=0.11\linewidth]{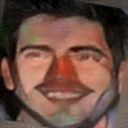}
\includegraphics[width=0.11\linewidth]{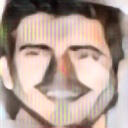}
\includegraphics[width=0.11\linewidth]{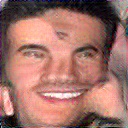}
\includegraphics[width=0.11\linewidth]{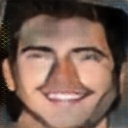}
\includegraphics[width=0.11\linewidth]{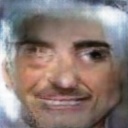}
\includegraphics[width=0.11\linewidth]{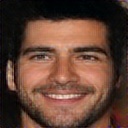}\\\vspace{0.1em}
\hspace{-3.9em}\includegraphics[width=0.11\linewidth]{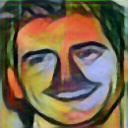}
\includegraphics[width=0.11\linewidth]{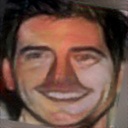}
\includegraphics[width=0.11\linewidth]{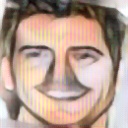}
\includegraphics[width=0.11\linewidth]{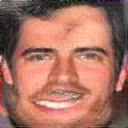}
\includegraphics[width=0.11\linewidth]{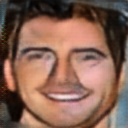}
\includegraphics[width=0.11\linewidth]{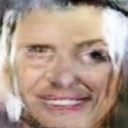}
\includegraphics[width=0.11\linewidth]{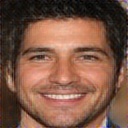}\\\vspace{0.1em}
\hspace{-3.9em}\includegraphics[width=0.11\linewidth]{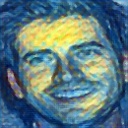}
\includegraphics[width=0.11\linewidth]{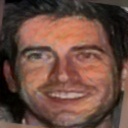}
\includegraphics[width=0.11\linewidth]{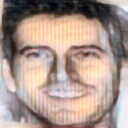}
\includegraphics[width=0.11\linewidth]{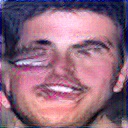}
\includegraphics[width=0.11\linewidth]{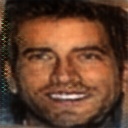}
\includegraphics[width=0.11\linewidth]{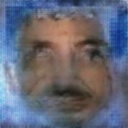}
\includegraphics[width=0.11\linewidth]{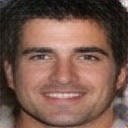}\\\vspace{-1.2mm}
\hspace{-3.9em}\subfigure[SF]{\label{fig:cmp4b}\scalebox{1}[1]{\includegraphics[width=0.11\linewidth]{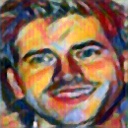}}}
\subfigure[\cite{gatys2016image}]{\label{fig:cmp4c}\scalebox{1}[1]
{\includegraphics[width=0.11\linewidth]{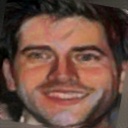}}}
\subfigure[\cite{johnson2016perceptual}]{\label{fig:cmp4d}\scalebox{1}[1]
{\includegraphics[width=0.11\linewidth]{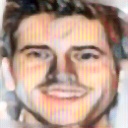}}}
\subfigure[\cite{li2016precomputed}]{\label{fig:cmp4e}\scalebox{1}[1]
{\includegraphics[width=0.11\linewidth]{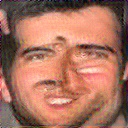}}}
\subfigure[\cite{isola2016image}]{\label{fig:cmp4f}\scalebox{1}[1]
{\includegraphics[width=0.11\linewidth]{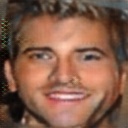}}}
\subfigure[\cite{zhu2017unpaired}]{\label{fig:cmp4g}\scalebox{1}[1]
{\includegraphics[width=0.11\linewidth]{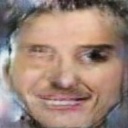}}}
\subfigure[Ours]{\label{fig:cmp4h}\scalebox{1}[1]
{\includegraphics[width=0.11\linewidth]{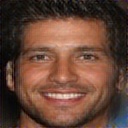}}}
\end{minipage}
\hfill
\vspace{-0.1cm}
\caption{Qualitative comparisons of the state-of-the-art methods. (a) The ground-truth real face. (b) Input portraits (from the test dataset) including the seen styles \emph{Candy, Feathers} and \emph{Scream} as well as the unseen styles \emph{Mosaic, Udnie, La Muse, Starry} and \emph{Composition VII}. (c) Gatys~\etal's method~\cite{gatys2016image}. (d) Johnson~\etal's method~\cite{johnson2016perceptual}. (e) Li and Wand's method~\cite{li2016precomputed} (MGAN). (f) Isola~\etal's method~\cite{isola2016image} (pix2pix). (g) Zhu~\etal's method~\cite{zhu2017unpaired} (CycleGAN). (h) Our method.}
\label{fig:cmp4}
\vspace{-0.3cm}
\end{figure*}

\begin{figure*}[!ht]
\hspace{4em}
\begin{minipage}{0.089\linewidth}
\centering
\subfigure[RF]{\label{fig:cmp5rf}\scalebox{1}[1]
{\includegraphics[width=1.15\linewidth]{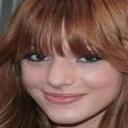}}} \hspace{0.2em}
\end{minipage}
\begin{minipage}{0.89\linewidth}
\centering
\hspace{-3.9em}\includegraphics[width=0.11\linewidth]{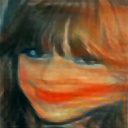}
\includegraphics[width=0.11\linewidth]{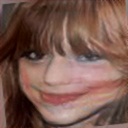}
\includegraphics[width=0.11\linewidth]{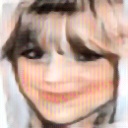}
\includegraphics[width=0.11\linewidth]{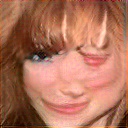}
\includegraphics[width=0.11\linewidth]{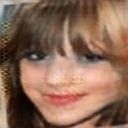}
\includegraphics[width=0.11\linewidth]{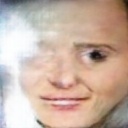}
\includegraphics[width=0.11\linewidth]{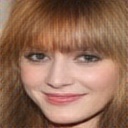}\\\vspace{0.1em}
\hspace{-3.9em}\includegraphics[width=0.11\linewidth]{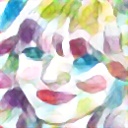}
\includegraphics[width=0.11\linewidth]{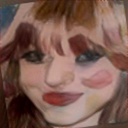}
\includegraphics[width=0.11\linewidth]{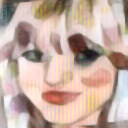}
\includegraphics[width=0.11\linewidth]{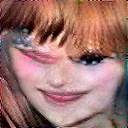}
\includegraphics[width=0.11\linewidth]{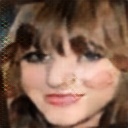}
\includegraphics[width=0.11\linewidth]{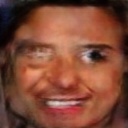}
\includegraphics[width=0.11\linewidth]{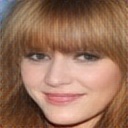}\\\vspace{0.1em}
\hspace{-3.9em}\includegraphics[width=0.11\linewidth]{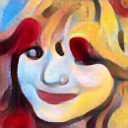}
\includegraphics[width=0.11\linewidth]{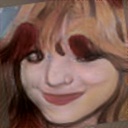}
\includegraphics[width=0.11\linewidth]{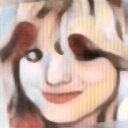}
\includegraphics[width=0.11\linewidth]{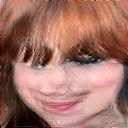}
\includegraphics[width=0.11\linewidth]{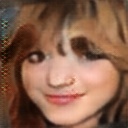}
\includegraphics[width=0.11\linewidth]{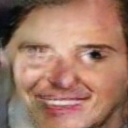}
\includegraphics[width=0.11\linewidth]{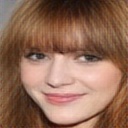}\\\vspace{0.1em}
\hspace{-3.9em}\includegraphics[width=0.11\linewidth]{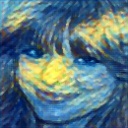}
\includegraphics[width=0.11\linewidth]{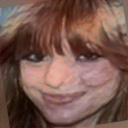}
\includegraphics[width=0.11\linewidth]{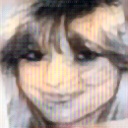}
\includegraphics[width=0.11\linewidth]{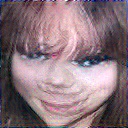}
\includegraphics[width=0.11\linewidth]{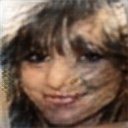}
\includegraphics[width=0.11\linewidth]{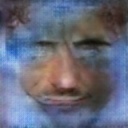}
\includegraphics[width=0.11\linewidth]{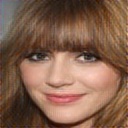}\\\vspace{0.1em}
\hspace{-3.9em}\includegraphics[width=0.11\linewidth]{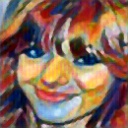}
\includegraphics[width=0.11\linewidth]{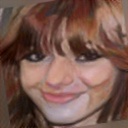}
\includegraphics[width=0.11\linewidth]{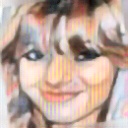}
\includegraphics[width=0.11\linewidth]{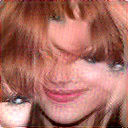}
\includegraphics[width=0.11\linewidth]{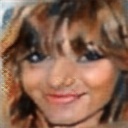}
\includegraphics[width=0.11\linewidth]{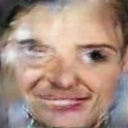}
\includegraphics[width=0.11\linewidth]{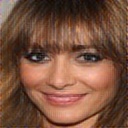}\\\vspace{0.1em}
\hspace{-3.9em}\includegraphics[width=0.11\linewidth]{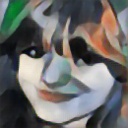}
\includegraphics[width=0.11\linewidth]{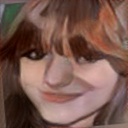}
\includegraphics[width=0.11\linewidth]{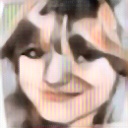}
\includegraphics[width=0.11\linewidth]{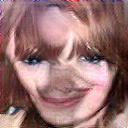}
\includegraphics[width=0.11\linewidth]{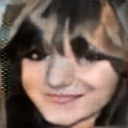}
\includegraphics[width=0.11\linewidth]{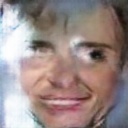}
\includegraphics[width=0.11\linewidth]{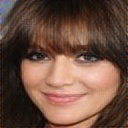}\\\vspace{0.1em}
\hspace{-3.9em}\includegraphics[width=0.11\linewidth]{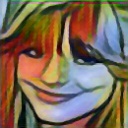}
\includegraphics[width=0.11\linewidth]{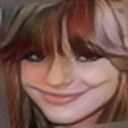}
\includegraphics[width=0.11\linewidth]{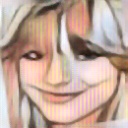}
\includegraphics[width=0.11\linewidth]{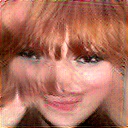}
\includegraphics[width=0.11\linewidth]{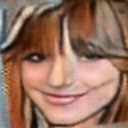}
\includegraphics[width=0.11\linewidth]{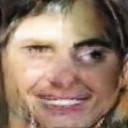}
\includegraphics[width=0.11\linewidth]{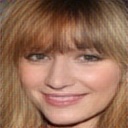}\\\vspace{-1.2mm}
\hspace{-3.9em}\subfigure[SF]{\label{fig:cmp5b}\scalebox{1}[1]{\includegraphics[width=0.11\linewidth]{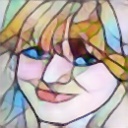}}}
\subfigure[\cite{gatys2016image}]{\label{fig:cmp5c}\scalebox{1}[1]
{\includegraphics[width=0.11\linewidth]{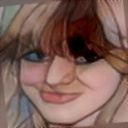}}}
\subfigure[\cite{johnson2016perceptual}]{\label{fig:cmp5d}\scalebox{1}[1]
{\includegraphics[width=0.11\linewidth]{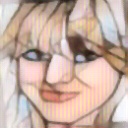}}}
\subfigure[\cite{li2016precomputed}]{\label{fig:cmp5e}\scalebox{1}[1]
{\includegraphics[width=0.11\linewidth]{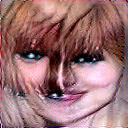}}}
\subfigure[\cite{isola2016image}]{\label{fig:cmp5f}\scalebox{1}[1]
{\includegraphics[width=0.11\linewidth]{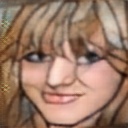}}}
\subfigure[\cite{zhu2017unpaired}]{\label{fig:cmp5g}\scalebox{1}[1]
{\includegraphics[width=0.11\linewidth]{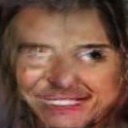}}}
\subfigure[Ours]{\label{fig:cmp5h}\scalebox{1}[1]
{\includegraphics[width=0.11\linewidth]{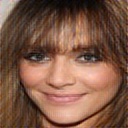}}}
\end{minipage}
\hfill
\vspace{-0.1cm}
\caption{Qualitative comparisons of the state-of-the-art methods. (a) The ground-truth real face. (b) Input portraits (from the test dataset) including the seen styles \emph{Scream, Feathers} and \emph{Candy} as well as the unseen styles \emph{Starry, Composition VII, Udnie, La Muse} and \emph{Mosaic}. (c) Gatys~\etal's method~\cite{gatys2016image}. (d) Johnson~\etal's method~\cite{johnson2016perceptual}. (e) Li and Wand's method~\cite{li2016precomputed} (MGAN). (f) Isola~\etal's method~\cite{isola2016image} (pix2pix). (g) Zhu~\etal's method~\cite{zhu2017unpaired} (CycleGAN). (h) Our method.}
\label{fig:cmp5}
\vspace{-0.3cm}
\end{figure*}

\noindent{\textbf{Visual Comparison with the state-of-art methods}}\\
Below, we provide several additional results demonstrating the performance of our IFRP network compared to the state-of-art approaches  (Figures \ref{fig:cmp3}, \ref{fig:cmp4} and \ref{fig:cmp5}).

\end{document}